%% file: arXiv.tex
\pdfoutput=1
\documentclass[final]{cvpr}

\usepackage{times}
\usepackage{epsfig}
\usepackage{graphicx}
\usepackage{amsmath}
\usepackage{amssymb}
\usepackage{cite}

\usepackage[utf8]{inputenc} 
\usepackage[T1]{fontenc}    
\usepackage{url}            
\usepackage{booktabs}       
\usepackage{amsfonts}       
\usepackage{nicefrac}       
\usepackage{microtype}      
\usepackage{xcolor}
\definecolor{citecolor}{HTML}{0071bc}
\usepackage{multirow}
\usepackage[title]{appendix}
\usepackage{xspace}
\usepackage{dsfont}
\usepackage{enumerate}
\usepackage{paralist}
\usepackage[normalem]{ulem}
\usepackage{chngcntr}

\makeatletter
\DeclareMathOperator*{\argmax}{argmax}

\DeclareRobustCommand\onedot{\futurelet\@let@token\@onedot}
\def\@onedot{\ifx\@let@token.\else.\null\fi\xspace}

\def\eg{\emph{e.g}\onedot} 
\def\ie{\emph{i.e}\onedot} 
 
\def\etc{\emph{etc}\onedot} 
 
\def\etal{\emph{et al}\onedot}

\makeatother

\usepackage[pagebackref=true,breaklinks=true,colorlinks=true,citecolor=citecolor,bookmarks=false]{hyperref}

\def\cvprPaperID{2062} 
\def\confYear{CVPR 2021}

\begin{document}

\title{A Realistic Evaluation of Semi-Supervised Learning \\
for Fine-Grained Classification}

\author{Jong-Chyi Su \quad \quad Zezhou Cheng \quad \quad  Subhransu Maji\\
University of Massachusetts Amherst\\
{\tt\small \{jcsu, zezhoucheng, smaji\}@cs.umass.edu}
}
\maketitle

\begin{abstract}
We evaluate the effectiveness of semi-supervised learning (SSL) on a realistic benchmark where data exhibits considerable class imbalance and contains images from novel classes.
Our benchmark consists of two fine-grained classification datasets obtained by sampling classes from the Aves and Fungi taxonomy.
We find that recently proposed SSL methods provide significant benefits, and can effectively use out-of-class data to improve performance when deep networks are trained from scratch.
Yet their performance pales in comparison to a transfer learning baseline, an alternative approach for learning from a few examples.
Furthermore, in the transfer setting, while existing SSL methods provide improvements, the presence of out-of-class is often detrimental.
In this setting, standard fine-tuning followed by distillation-based self-training is the most robust.
Our work suggests that semi-supervised learning with experts on realistic datasets may require different strategies than those currently prevalent in the literature.
\end{abstract}

\input{intro}

\input{related}
\input{dataset}

\input{methods}

\input{experiments}
\input{conclusion}

{\small{\paragraph{Acknowledgements} This project is supported in part by NSF \#1749833 and was performed using high performance computing equipment obtained under a grant from the Collaborative R\&D Fund managed by the Massachusetts Technology Collaborative.}}

\appendix
\input{supp_arXiv}

{\small
\bibliographystyle{ieee_fullname}
\bibliography{egbib,reference,reference_ssl}
}

\end{document}

%% file: intro.tex
\section{Introduction}

\begin{figure}
\centering
\includegraphics[width=0.49\linewidth]{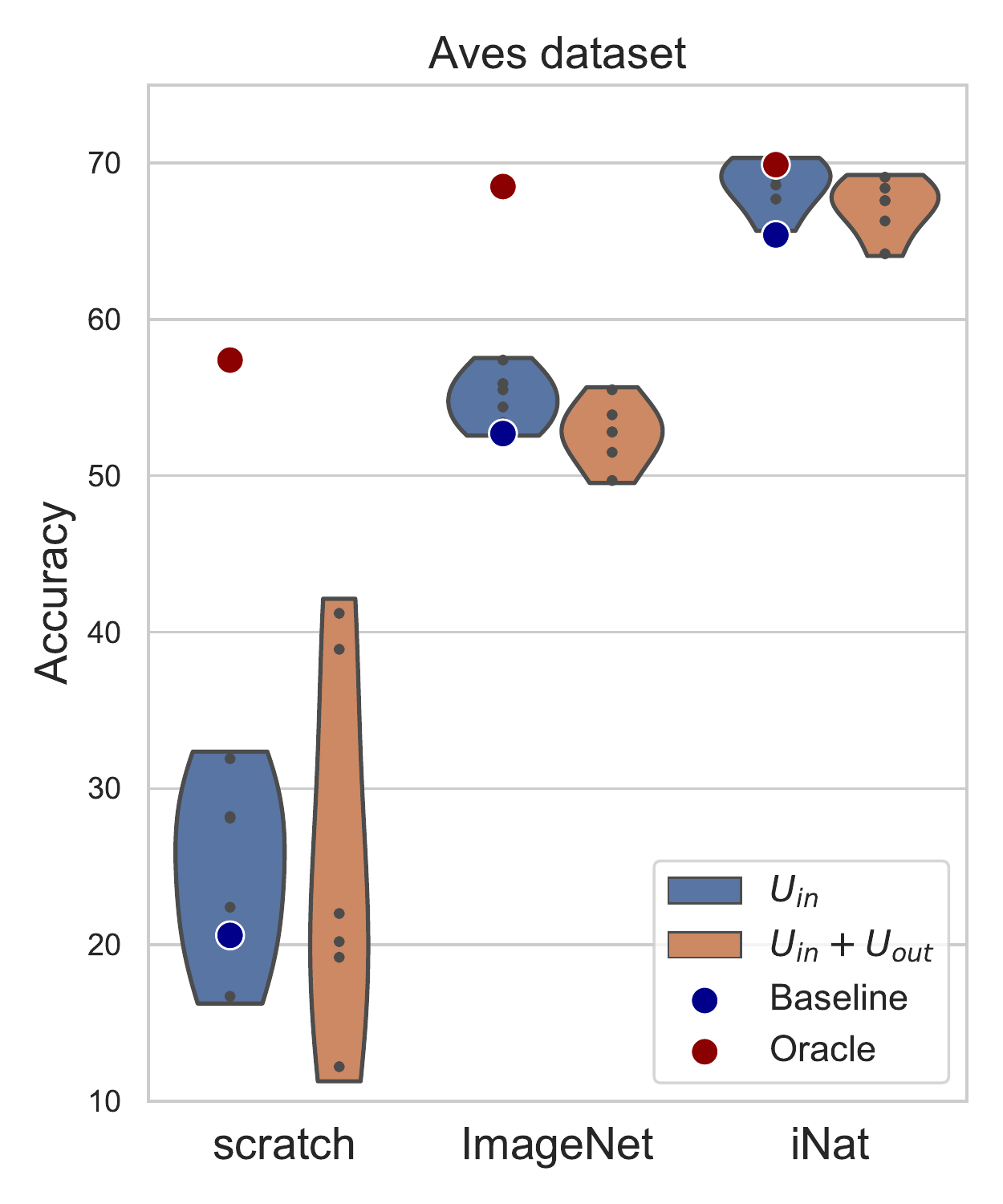}
\includegraphics[width=0.49\linewidth]{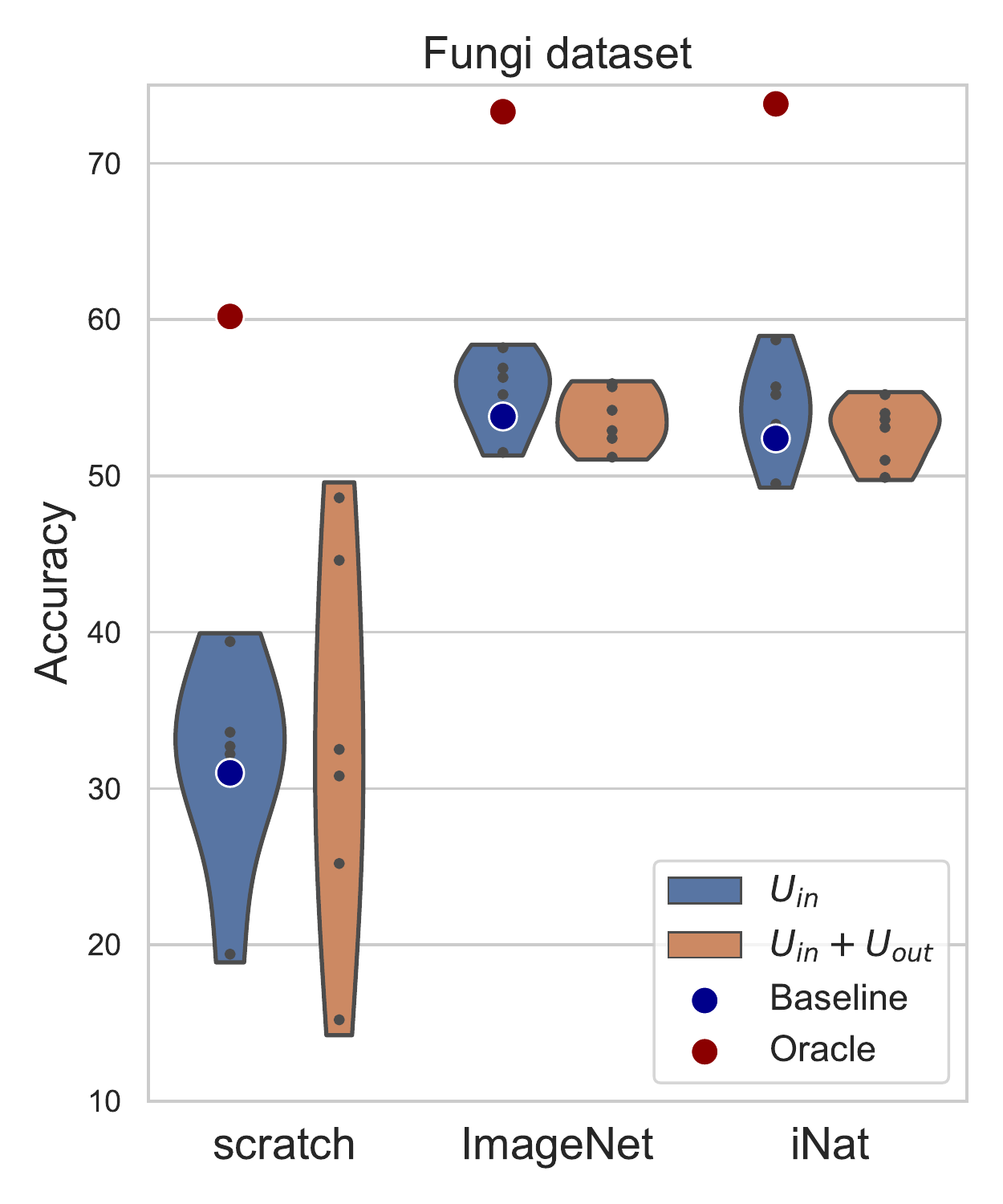}
\caption{Accuracy of semi-supervised learning (SSL) algorithms on the Semi-Aves and Semi-Fungi datasets (see Fig.~\ref{fig:dataset}) using (i) different pre-trained models, and (ii) in-class ($U_{in}$) and out-of-class ($U_{in} + U_{out}$) unlabeled data. The performances of the supervised baseline and supervised oracle are also shown. Transfer learning from experts is far more effective than SSL from \emph{scratch}, while in the transfer setting SSL provides modest gains. Though out-of-class data ($U_{out})$ is valuable when training from scratch, it is not the case when training from experts (details in Tab.~\ref{tab:benchmark_aves}~and~\ref{tab:benchmark_fungi}).}
\label{fig:violin}
\end{figure}

\begin{figure*}[t!]
\centering
\includegraphics[width=\linewidth]{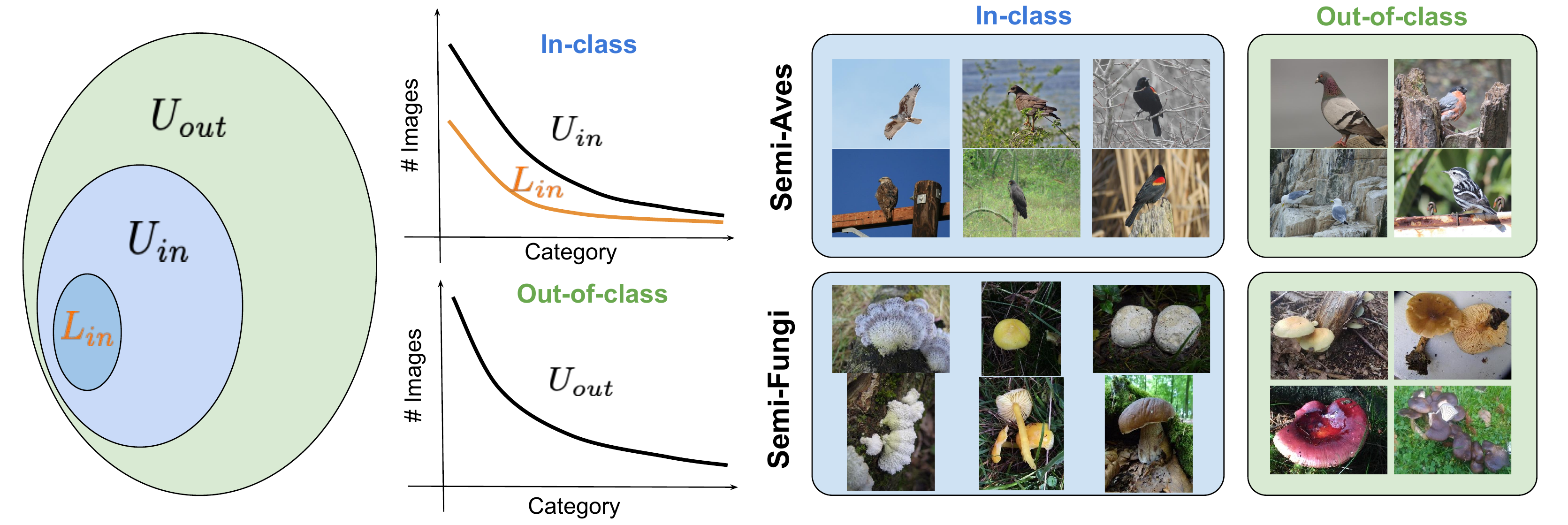}
\caption{\textbf{The proposed benchmark for semi-supervised learning.} The benchmark contains two datasets, with classes from the Aves and Fungi taxa respectively. Each represents a 200-way classification task and the training set contains (i) labeled images from these classes $L_{in}$, (ii) unlabeled images from these classes $U_{in}$, and (iii) unlabeled images from related classes $U_{out}$, as seen on figures to the right. Moreover, the classes exhibit a long-tailed distribution with an imbalance ratio of 8 to 10. The benchmark captures conditions observed in some realistic applications that are not present in existing datasets used to evaluate semi-supervised learning. See \S~\ref{sec:dataset} and Tab.~\ref{tab:dataset} for details.}

\label{fig:dataset}
\end{figure*}
Semi-supervised learning (SSL) aims to exploit unlabeled data to train models from a few labels, making them practical for applications where labels are a bottleneck.
Yet, the current literature on SSL with deep networks for image classification has two main shortcomings. 
First, most methods are evaluated on curated datasets such as CIFAR, SVHN, or ImageNet, where class distribution is or is close to uniform and unlabeled data contains no novel classes.
This is implicit in methods that rely on the assumption that the data is uniformly clustered,
use a uniform instead of class-balanced loss, or categorize unlabeled data into one of the labeled classes.
In practice, however, class distribution can be highly unbalanced or even unknown, and the unlabeled data may contain novel classes. How effective is SSL in these situations?

Second, most literature has focused on training models from scratch. However, a practical approach for few-shot learning is to use expert models trained on large labeled datasets such as ImageNet~\cite{ILSVRC15} or iNaturalist~\cite{van2018inaturalist}.
What gains does SSL provide in this setting, especially since many SSL methods are based on learning invariances from data based on transformations which might have already been learned by the experts during supervised training?
Moreover, is out-of-domain data beneficial when experts are available?

Our paper aims to answer these questions by conducting a systematic study of SSL techniques (Fig.~\ref{fig:violin}) on two fine-grained classification datasets that exhibit a long-tailed distribution of classes and contain a large number of out-of-class images (Fig.~\ref{fig:dataset}).
These datasets are obtained by sampling classes under the Aves (birds) and Fungi taxonomy. The out-of-class images are other Aves (or Fungi) images not belonging to the classes within the labeled set. 
The first dataset was part of the semi-supervised challenge at FGVC7 workshop~\cite{su2021semisupervised}, while the second one is constructed from the FGVC fungi challenge~\cite{fungi_challenge} following a similar scheme, details of which are described in \S~\ref{sec:dataset}. 
We also provide a benchmark on the CUB dataset~\cite{WahCUB_200_2011} in the appendix.

On these datasets, we conduct a systematic study of existing deep-learning-based semi-supervised learning approaches for image classification.
We perform experiments on SSL methods including Pseudo-Label~\cite{lee2013pseudo}, Curriculum Pseudo-Label~\cite{cascante2020curriculum}, FixMatch~\cite{sohn2020fixmatch}, self-training using distillation~\cite{xie2020self}, self-supervised learning (MoCo~\cite{he2020momentum}), as well as their combinations when effective.
We investigate strategies for using unlabeled data when models are initialized from experts.
We also evaluate the performance of methods that use unlabeled data from the same classes as the labeled dataset ($U_{in}$) and a practical setting where the unlabeled data includes out-of-class images ($U_{in} + U_{out}$). The high-level summary of the experiments reported in Fig.~\ref{fig:violin}, Tab.~\ref{tab:benchmark_aves}, \ref{tab:benchmark_fungi}, and Fig.~\ref{fig:relative} are as follows:

\setdefaultleftmargin{1em}{2em}{}{}{}{}
\begin{compactitem}
  \setlength\itemsep{0.5em}
    \item Some of the SSL methods are effective when models are trained from scratch, especially those with self-supervised pre-training can significantly benefit from out-of-class data (long blue whiskers and longer orange whiskers above the \emph{baseline} for \emph{scratch} in Fig.~\ref{fig:violin}). In this setting, self-supervised learning followed by distillation-based self-training performs the best (Tab.~\ref{tab:benchmark_aves} and \ref{tab:benchmark_fungi}).
    
    \item The best SSL approach significantly under-performs the supervised fine-tuning model trained on the labeled portion of the datasets (the \emph{baseline} performance of \emph{ImageNet} and \emph{iNat} is higher than any SSL model trained from scratch in Fig.~\ref{fig:violin}).
    
    \item Picking the right expert provides further gains in this few-shot setting but not when training using the entire labeled dataset (\emph{oracle} performance in Fig.~\ref{fig:violin}).
    
    \item When training with experts, FixMatch gives the most improvements when having $U_{in}$ only.
    However, the presence of out-of-class unlabeled data often hurts performance.
    Self-Training was the most robust to the presence of out-of-class data (Tab.~\ref{tab:benchmark_aves}, \ref{tab:benchmark_fungi} and Fig.~\ref{fig:relative}).
    
    \item Surprisingly, we found that no method was able to reliably use out-of-class data even though the domain shift is relatively small (the orange group is not higher than the blue groups for \emph{ImageNet} and \emph{iNat} unlike \emph{scratch} in Fig.~\ref{fig:violin}), echoing the experience of participants in the FGVC7 challenge~\cite{su2021semisupervised}.
    
    \item The performance of SSL is far below the model trained using labels of the entire in-class data suggesting that there is significant room for improvement (\emph{oracle} performance in Fig.~\ref{fig:violin}).
\end{compactitem}

In summary, we conduct a systematic evaluation of several recently proposed SSL techniques on two challenging datasets representing a long-tailed distribution of fine-grained categories. We vary the initialization and the domain of the unlabeled data and analyze the robustness of various SSL approaches. 
Our experiments indicate that SSL does not work out-of-the-box in a transfer learning setting, 
especially in the presence of out-of-domain data. 
These results are in a similar vein to prior work on the evaluation of SSL approaches that have analyzed the robustness of SSL techniques to the choice of hyper-parameters~\cite{oliver2018realistic}, network architectures~\cite{chen2020big,xie2019unsupervised}, and domain shifts~\cite{oliver2018realistic,Su2020When,Wallace2020Extending}, \etc.
However, the evaluation in a transfer learning setting on the proposed benchmarks reveals additional insights.
We hope these experiments inspire practical methods that combine the benefits of supervised learning and task-specific learning on partially labeled datasets.

%% file: related.tex
\section{Related Works}
Semi-supervised learning has a long history in machine learning. In this section, we describe the trends in recent techniques based on deep learning and refer the reader to surveys on SSL for a comprehensive view~\cite{chapelle2009semi,zhu2009introduction,zhu2005semi,van2020survey}. 

\paragraph{Self-Training.} These techniques use the model's prediction to automatically generate labels for the unlabeled data~\cite{mclachlan1975iterative,scudder1965probability}. 
Pseudo-Labeling~\cite{lee2013pseudo} includes confident predictions, \ie, those greater than a threshold for training. 
The pseudo-labels can be added iteratively to induce a ``curriculum''~\cite{cascante2020curriculum,bengio2009curriculum,hacohen2019power}.
Alternatively, one can add an entropy penalty to encourage confident predictions on the unlabeled data~\cite{grandvalet2005semi}.
Other methods~\cite{xie2020self,yalniz2019billion,zoph2020rethinking,chen2020big} involve re-training a ``student model'' from a ``teacher model'' using its prediction computed in different ways.
For example, adding noise and using a larger student model~\cite{xie2020self, zoph2020rethinking}, selecting k-most confident pseudo-labels~\cite{yalniz2019billion}, or using a distillation loss which softens the predictions~\cite{xie2020self,chen2020big}.
While these methods have been shown to be successful in various datasets, the effectiveness of the approach is critically dependent on the initial performance of the model and the data distribution. Our experiments show that the presence of out-of-class data negatively impacts some of these methods while using expert initialization provides a significant benefit.

\paragraph{Consistency-based learning.} 
These methods learn by encouraging the consistency of the model's predictions on the unlabeled data.
These could be across different augmentations of the data~\cite{bachman2014learning,rasmus2015semi,laine2016temporal,sajjadi2016regularization}, including adversarial versions~\cite{miyato2018virtual}.
Alternatively, consistency can be enforced across time, \eg, using moving average of the predictions (\emph{temporal ensembling}~\cite{laine2016temporal}), using the moving average of model parameters (\emph{mean teacher}~\cite{tarvainenweight}), or using a stochastic averaging of model parameters~\cite{athiwaratkun2018there}.
A number of methods for \emph{data augmentation} have been proposed which has generally improved both supervised and semi-supervised learning. These include the variety of image augmentations proposed in RandAugment~\cite{cubuk2020randaugment}, the CutOut scheme~\cite{devries2017improved}, linear combinations of images used in MixUp~\cite{zhang2017mixup}, and even augmentations in the feature space~\cite{kuo2020featmatch}.
These augmentations have been incorporated in methods such as MixMatch~\cite{berthelot2019mixmatch}, ReMixMatch~\cite{berthelot2019remixmatch}, FixMatch~\cite{sohn2020fixmatch}, UDA~\cite{xie2019unsupervised}, and ICT~\cite{verma2019interpolation} in different ways for consistency-based learning. 
We choose FixMatch as the candidate approach which has shown state-of-the-art results on existing SSL benchmarks, which we describe in detail in \S~\ref{sec:methods}.
While consistency via data-augmentation is effective when a model is trained from scratch, it is unclear if this is effective when using a pre-trained model, as invariance to these transformations may have been acquired during supervised pre-training. 

\paragraph{Self-supervised learning.} Another line of work has explored using self-supervised (or unsupervised) learning objectives to improve semi-supervised learning.
These include incorporating pre-text tasks such as predicting image rotations~\cite{gidaris2018unsupervised}, the order of patches (jigsaw puzzle task)~\cite{noroozi2016unsupervised} during semi-supervised learning~\cite{zhai2019s, Su2020When,rebuffi2020semi}.
Alternatively, self-supervised learning can be used as an initialization before training with labels. 
The recent success of self-supervised learning based on contrastive learning~\cite{oord2018representation,hjelm2018learning,he2020momentum,tian2019contrastive,chen2020big} has been incorporated by several approaches leading to promising results on ImageNet~\cite{chen2020big}. 
We also find that contrastive learning followed by self-training is the best performing when trained from scratch on our benchmark. However, the value of out-of-class data is diminished when using expert models.

\paragraph{Analysis of semi-supervised learning.} The most related to our work is that of Oliver~\etal~\cite{oliver2018realistic} who provide a benchmark for comparing deep-learning-based SSL methods for image classification. 
While only CIFAR10 and SVHN datasets were used, the paper pointed out that hyper-parameters can have a significant impact on performance. Yet these are hard to tune without access to large amounts of labeled data, which is precisely the setting in semi-supervised learning. In our analysis, we pay careful attention to hyper-parameter optimization (see \S~\ref{sec:implementation}). 
Their work also showed that transfer learning from experts can be more effective, but did not explore if their combination with semi-supervised learning can be helpful. 
In addition, they showed out-of-class unlabeled data may be harmful, but the classes in $U_{in}$ and $U_{out}$ are widely different.
Last, their work as well as most SSL methods have been presented on well-curated datasets.
Our work focuses on the evaluation in a realistic setting and includes an analysis of methods proposed since Oliver~\etal's paper.
These include methods based on self-training, self-supervised training, and FixMatch which outperform the consistency-based approach~\cite{miyato2018virtual} analyzed in their paper.
More comparisons between~\cite{oliver2018realistic} are provided in the appendix.

%% file: dataset.tex
\section{A Realistic Benchmark}\label{sec:dataset}

\begin{table*}[tbp]
  \setlength{\tabcolsep}{7.5pt}
  \renewcommand{\arraystretch}{1.1}
  \centering
  \begin{tabular}{c c c c c c c}
    \toprule
    \multirow{2}{*}{\textbf{Dataset}} & \textbf{Classes} & \textbf{Images} & \textbf{Unlabeled} & \textbf{Image} & \textbf{Class} & \textbf{Imbalance}\\
     & ${L}_{in}$ / ${U}_{in}$ / ${U}_{out}$ & ${L}_{in}$ / ${U}_{in}$ / ${U}_{out}$ & \textbf{Class Domain} & \textbf{Resolution} & \textbf{Distribution} & \textbf{Ratio}\\
    \midrule
    CIFAR-10 & 10 / 10 / 0 & 4K / 40K / 0 & ${L}={U}$& 32$\times$32 & uniform & 1\\
    CIFAR-100 & 100 / 100 / 0 & 10K / 50K / 0 & ${L}={U}$& 32$\times$32 & uniform & 1\\
    SVHN & 10 / 10 / 0 & 1K / 65K / 0 & ${L}={U}$& 32$\times$32 & uniform & 1\\
    STL-10 & 10 / 0 / - & 5K / 0 / 100K & ${L}\neq{U}$& 96$\times$96 & uniform & 1\\
    ImageNet & 1000 / 1000 / 0 & 140K / 1.26M / 0 & ${L}={U}$& 224$\times$224 & $\approx$ uniform & 1.8\\
    Semi-Aves & 200 / 200 / 800 & 6K / 27K / 122K & ${L}={U}_{in}\neq{U}_{out}$& 224$\times$224 & long-tailed & 7.9\\
    Semi-Fungi & 200 / 200 / 1194 & 4K / 13K / 65K & ${L}={U}_{in}\neq{U}_{out}$& 224$\times$224 & long-tailed & 10.1\\
    \bottomrule
  \end{tabular}
  \caption{A comparison of Semi-Aves and Semi-Fungi datasets with existing SSL benchmarks. The Semi-Aves and Semi-Fungi present a challenge due to the large number of classes, presence of novel images in the unlabeled set, long-tailed distribution of classes as indicated by the class imbalance ratio (maximum / minimum images per class) in the training set.}
  \label{tab:dataset}
\end{table*}

In SSL, we are provided with labeled training data $(x_i,y_i) \in {L}_{in}$ and unlabeled training data $(u_i,\cdot) \in {U}$. 
The unlabeled data can either belong to the same classes as the labeled data ($U_{in}$), or to novel classes ($U_{out}$).
In a realistic setting, one may expect that the unlabeled data contains novel classes. 
In many applications it is easy to acquire images from related domains through coarse labeling, \eg, it is easier to label an image as a bird than a ``yellow bunting''.
Such images could be potentially used to learn better representations.
Thus we evaluate SSL methods in two settings, one when the unlabeled data contains no novel images, and another when it does, \ie, $U_{in}$ and $U_{in} + U_{out}$ respectively.

We use two datasets by sampling classes from the natural domains for our benchmark. 
As shown in Fig.~\ref{fig:dataset}, the classes belong to the Aves and Fungi taxonomy and contain a long-tailed distribution of classes, as commonly observed in fine-grained domains. 
Tab.~\ref{tab:dataset} shows a comparison with other benchmarks. 
Larger image sizes, significant class imbalance, fine-grained categories, and a large number of out-of-class images allow a more realistic evaluation of SSL techniques. Below we describe each dataset.

\paragraph{Semi-Aves.}
We use the dataset from the semi-supervised challenge at the FGVC7 workshop at CVPR 2020~\cite{su2021semisupervised}. The dataset includes a subset of bird species from the Aves kingdom of iNaturalist 2018 dataset~\cite{van2018inaturalist}. 
However, there are no overlapping images since the images were collected from recent years. 
There are 200 in-class and 800 out-of-class categories. 
The training and validation set has a total of 5959 labeled images, 26,640 and 122,208 in-class and out-of-class unlabeled images, and 8000 test images. 
The training data in $L_{in}$, $U_{in}$, and $U_{out}$ is long-tail distributed, specifically $L_{in}$ has 15 to 53 images and $U_{in}$ has 16 to 229 images per class.
The test data has a uniform distribution with 40 images per class. 

\paragraph{Semi-Fungi.} We create a Semi-Fungi dataset following the similar strategy of the Semi-Aves dataset. 
We use the train-val set of images from the FGVCx Fungi challenge at the FGVC5 workshop at CVPR 2018~\cite{fungi_challenge}. 
The dataset was collected from the ``Svampe Atlas''\footnote{\url{https://svampe.databasen.org}} website, thus the image domain is different from iNaturalist.
The original dataset has 1394 fungi species with a long-tailed distribution. 
We first sort the classes by frequency and randomly select 200 of the top 600 classes as in-class categories. 
We then select 20 images per class as the test set, and randomly select 4141 images as labeled data and the rest 13,166 images as in-class unlabeled data. 
The rest 1194 species are used as out-of-class unlabeled images, which has a total of 64,871 images.
In Semi-Fungi, there are 6 to 78 images per class in $L_{in}$, and 16 to 276 images in $U_{in}$. The test set is uniformly distributed with 20 images per class.

%% file: methods.tex
\section{Methods}\label{sec:methods}
In this section, we describe the details of the SSL methods we compared in our benchmark.
{\flushleft {\bf (1) Supervised baseline / oracle:}} 
We train the model only using labeled data $L_{in}$ with a cross-entropy loss. 
For the oracle, we include the ground-truth labels of $U_{in}$ for training. 

{\flushleft {\bf (2) Pseudo-Labeling~\cite{lee2013pseudo}:}} 
The approach uses a base model's confident predictions on unlabeled images as labels. Concretely, if the maximum probability of a class is greater than a threshold $\tau$, we then take the class as the target label.
Following the implementation of Oliver \etal~\cite{oliver2018realistic}, we sample half of the batch from $L_{in}$ and half from unlabeled data $U$ during training. 
Denote $(x_i, y_i)$ as a labeled sample, the predictions on unlabeled data $u_i$ of the model $f$ as $q_i = f(u_i)$, pseudo-label as $\hat{q}_i = \argmax(q_i)$, and  cross-entropy function as $H(p,q)=-\sum_{r}p(r)\log q(r)$.
Then, the objective for each batch is:
\begin{equation}
\mathcal{L} = \sum_{j=1}^{n} H (y_i, f(x_i)) + \sum_{i=1}^{n}\mathds{1}\big[\max(q_i) \geq \tau\big] H (\hat{q_i},q_i) .
\end{equation}

{\flushleft {\bf (3) Curriculum Pseudo-Labeling~\cite{cascante2020curriculum}:}} 
Unlike pseudo-labeling where labels are generated in an online manner, curriculum labeling generates pseudo-labels after the training is finished on the current labeled set before retraining.
We first train a supervised model on labeled data $L_{in}$, then select images with the highest predictions from all the unlabeled data $u \in {U}$, and add them with their pseudo-labels to the labeled dataset. In the next iteration, we retrain a model \emph{from scratch} using the new set of labeled data. We repeat this process 5 times and select \{20, 40, 60, 80, 100\}\% of the unlabeled data from the original pool of unlabeled data $U$. The steps are as the following:
\begin{enumerate}[(i)]
\setlength\itemsep{0.0em}
\item Initialize $L=L_{in}$, $\beta = 20$.
\item Supervised training on $L$ .
\item Generate predictions $q=f_\theta(x)$ for every $u \in {U}$. 
\item From ${U}$ select $\beta$\% examples with highest prediction scores and their pseudo-labels as ${L}_{top}$.
\item Add selected unlabeled data with their pseudo-labels to the labeled dataset ${L} = {L_{in}} \cup {L}_{top}$. 
\item If $\beta<$100, $\beta=\beta+$20 and repeat from step (ii) . 
\end{enumerate}

{\flushleft {\bf (4) FixMatch:}} 
FixMatch combines pseudo-labeling and consistency regularization.
For each unlabeled image, it minimizes the cross-entropy between the pseudo-label (thresholded prediction) of the weakly-augmented image and the predictions of the strong-augmented image. 
For labeled data, only weak augmentations are applied.
Specifically, let $\alpha$ be a weak augmentation (image flipping in our case) and $\mathcal{A}$ be a strong augmentation (RandAugment~\cite{cubuk2020randaugment} in our case). Let the predictions under strong and weak augmentations are $Q_i = f(\mathcal{A}(u_i))$, $q_i = f(\alpha(u_i))$. 
The total loss for labeled and unlabeled data is
\begin{equation}\label{eq:fixmatch}
\mathcal{L} = \sum_{j=1}^{m} H (y_j,f(\alpha(x_j))) + \sum_{i=1}^{k m }\mathds{1}\big[\max(q_i) \geq \tau \big] H (\hat{q_i},Q_i).
\end{equation}
In the original implementation, each batch uses $m$ labeled and $k m$ unlabeled data with a total batch size $n=(k+1)m$, where the sampling ratio $k$ is a hyper-parameter. 

{\flushleft {\bf (5) Self-Training:}} 
While the term of ``Self-Training'' is general, we use this to refer to the following procedure using distillation~\cite{hinton2015distilling}.
We first train a supervised model $f^t$ on the labeled data which we call the teacher model, then train a student model $f^s$ with scaled cross-entropy loss on the unlabeled data and cross-entropy loss on labeled data.
Distillation was originally used for model compression~\cite{bucilua2006model}, but has been shown to improve the performance when training the student model with the same architecture~\cite{furlanello2018born} or across different modalities~\cite{tian2019contrastive,gupta2016cross,su2016adapting}. 
Given unlabeled data $(u,\cdot)$, let the \emph{logits} from teacher and student model as $z^t$ and $z^s$, and the prediction of labeled data $(x,y)$ from the student model is $y^s$. The objective includes the cross-entropy loss for labeled data $(x,y)$, and the distillation loss for unlabeled data:
\begin{equation}\label{eq:distillation}
\mathcal{L} = (1-\lambda)\sum_{j=1}^{n}H(y_j,y_j^s) + \lambda \sum_{i=1}^{n}H \left(\sigma\left(\frac{z_i^t}{T}\right), \sigma\left(\frac{z_i^s}{T}\right)\right),
\end{equation}
where $\lambda$ is the weight between supervised and distillation losses, $\sigma$ is the softmax function, and $T$ is a temperature (scaling) parameter. 

{\flushleft {\bf (6) Self-Supervised Learning (MoCo~\cite{he2020momentum}):}} 
We use momentum contrastive (MoCo) learning as a strong baseline for self-supervised training. MoCo learns an image encoder $f(x)$ that maps the image $x$ to a representation $q = f(x)$ and uses a contrastive objective that requires positive pairs to be closer than negative pairs in the representation space. The positive pairs are sampled from two geometric or photometric augmented views of a same images while negative images are augmentations from different images. MoCo adapts the InfoNCE~\cite{oord2018representation} loss as the objective function. The loss for each encoded query $q$ is: 
\begin{equation}\label{eq:moco}
\mathcal{L}_q = -\log\frac{\exp\left( q \cdot k^{+} / T \right)}{\exp( q \cdot k^{+} / T) + \sum_i^K \exp(q \cdot k_i^{-} / T )},
\end{equation}
where $T$ is the temperature, $k^{+}$ and $k^-$ are the positive and negative sample of the query $q$. 
The number of negative samples $K$ is limited by the mini-batch size.
In order to stabilize the training, MoCo uses the memory bank~\cite{wu2018unsupervised} to store the negative samples and updates the encoder of the keys in the memory bank based on momentum.
After the self-supervised pre-training, we remove the MLP layers after the global average pooling layer, add a linear classifier (a fully convolutional layer followed by softmax), and train the entire network with supervised cross-entropy loss. We found that freezing the pre-trained backbone gives worse performance than fine-tuning the entire network.

{\flushleft {\bf (7) MoCo + Self-Training:}} 
Here we initialize the model using MoCo learning on the unlabeled data before semi-supervised learning using Self-Training. 
A recent work by Chen~\etal~\cite{chen2020big} has shown this to be a strong semi-supervised learning baseline. The procedure is as follows:
\begin{enumerate}[(i)]
\setlength\itemsep{0.0em}
\item Pre-train the model using MoCo on $L_{in}$ and $U$.
\item Fine-tune the model on $L_{in}$ with a cross-entropy loss. Call this the teacher model $f^t$.
\item Train a student model $f^s$ initialized from step (i) with distillation loss (Eq.~\ref{eq:distillation}) using the teacher model $f^t$.
\end{enumerate}

%% file: experiments.tex
\section{Experiments}
\definecolor{mygray}{gray}{0.7}
\newcommand\grey[1]{\textcolor{mygray}{#1}}
\newcommand\best[1]{\textbf{\textcolor{teal}{#1}}}
\begin{table*}[tbp]
  \setlength{\tabcolsep}{7pt}
  \renewcommand{\arraystretch}{1.1}
  \centering
  \begin{tabular}{c c | c c | c c | c c}
    \toprule
    &\multirow{2}{*}{Method} & \multicolumn{2}{c|}{\textbf{from scratch}} & \multicolumn{2}{c|}{\textbf{from ImageNet}} & \multicolumn{2}{c}{\textbf{from iNat}}\\
    && Top1 & Top5 & Top1 & Top5 & Top1 & Top5 \\
    \midrule
    & Supervised baseline & 20.6$\pm$0.4 & 41.7$\pm$0.7 & 52.7$\pm$0.2 & 78.1$\pm$0.1 & 65.4$\pm$0.4 & 86.6$\pm$0.2\\
    &\grey{Supervised oracle} & \grey{57.4$\pm$0.3} & \grey{79.2$\pm$0.1} & \grey{68.5$\pm$1.4} & \grey{88.5$\pm$0.4} & \grey{69.9$\pm$0.5} & \grey{89.8$\pm$0.7}\\
    \midrule
    \multirow{6}{*}{\rotatebox[origin=c]{90}{${U}_{in}$}} & Pseudo-Label~\cite{lee2013pseudo} & 16.7$\pm$0.2 & 36.5$\pm$0.8 & 54.4$\pm$0.3 & 78.8$\pm$0.3 & 65.8$\pm$0.2 & 86.5$\pm$0.2\\
    & Curriculum Pseudo-Label~\cite{cascante2020curriculum} & 20.5$\pm$0.5 & 41.7$\pm$0.5 & 53.4$\pm$0.8 & 78.3$\pm$0.5 & {69.1$\pm$0.3} & {87.8$\pm$0.1}\\
    & FixMatch~\cite{sohn2020fixmatch} & 28.1$\pm$0.1 & 51.8$\pm$0.6 & \best{57.4$\pm$0.8}  & {78.5$\pm$0.5}  & \best{70.2$\pm$0.6} & 87.0$\pm$0.1 \\
    & Self-Training & 22.4$\pm$0.4 & 44.1$\pm$0.1 & {55.5$\pm$0.1} & {79.8$\pm$0.1} & 67.7$\pm$0.2 & 87.5$\pm$0.2\\
    & MoCo ~\cite{he2020momentum} & 28.2$\pm$0.3 & 53.0$\pm$0.1 & 52.7$\pm$0.1 & 78.7$\pm$0.2 & 68.6$\pm$0.1 & 87.7$\pm$0.1\\
    & MoCo + Self-Training & \best{31.9$\pm$0.1} & \best{56.8$\pm$0.1} & {55.9$\pm$0.2} & \best{80.3$\pm$0.1} & \best{70.1$\pm$0.2} & \best{88.1$\pm$0.1}\\
    \midrule
    \multirow{6}{*}{\rotatebox[origin=c]{90}{${U}_{in}+{U}_{out}$}} & Pseudo-Label~\cite{lee2013pseudo} & 12.2$\pm$0.8 & 31.9$\pm$1.6 & 52.8$\pm$0.5 & 77.8$\pm$0.1 & 66.3$\pm$0.3 & 86.4$\pm$0.2 \\
    & Curriculum Pseudo-Label~\cite{cascante2020curriculum} & 20.2$\pm$0.5 & 41.0$\pm$0.9 & 52.8$\pm$0.5 & 77.8$\pm$0.1 & \best{69.1$\pm$0.1} & \best{87.6$\pm$0.1}\\
    & FixMatch~\cite{sohn2020fixmatch} & 19.2$\pm$0.2 & 42.6$\pm$0.6 & 49.7$\pm$0.2 & 72.8$\pm$0.5 & 64.2$\pm$0.2 & 84.5$\pm$0.1\\
    & Self-Training & 22.0$\pm$0.5 & 43.3$\pm$0.2 & \best{55.5$\pm$0.3} & \best{79.7$\pm$0.2} & {67.6$\pm$0.2} & \best{87.6$\pm$0.1}\\
    & MoCo~\cite{he2020momentum} & 38.9$\pm$0.4 & 65.4$\pm$0.3 & 51.5$\pm$0.4 & 77.9$\pm$0.2 & {67.6}$\pm$0.1 & \best{87.3$\pm$0.2}\\
    & MoCo + Self-Training & \best{41.2$\pm$0.2} & \best{65.9$\pm$0.3} & {53.9$\pm$0.2} & \best{79.4$\pm$0.3} & {68.4$\pm$0.2} & \best{87.6$\pm$0.2}\\
    \bottomrule
  \end{tabular}
  \caption{\textbf{Results on Semi-Aves benchmark.} We experiment with six different SSL methods as well as supervised baselines under different settings: (1) using $U_{in}$ or $U_{in} + U_{out}$ as the unlabeled dataset, (2) training from scratch, or using ImageNet or iNat pre-trained model. We show that when training from scratch with $U_{in}$, MoCo + Self-Training performs the best. When having expert models, transfer learning is a strong baseline, and FixMatch and Self-Training can still give improvements. When adding unlabeled data from $U_{out}$, the performance pales except for the self-supervised method when training from scratch. 
  The best results and those within the variance are marked in \best{teal}.
  }
  \label{tab:benchmark_aves}
\end{table*}

\begin{table*}[tbp]
  \setlength{\tabcolsep}{7pt}
  \renewcommand{\arraystretch}{1.1}
  \centering
  \begin{tabular}{c c | c c | c c | c c}
    \toprule
    &\multirow{2}{*}{Method} & \multicolumn{2}{c|}{\textbf{from scratch}} & \multicolumn{2}{c|}{\textbf{from ImageNet}} & \multicolumn{2}{c}{\textbf{from iNat}}\\
    && Top1 & Top5 & Top1 & Top5 & Top1 & Top5 \\
    \midrule
    & Supervised baseline & 31.0$\pm$0.4 & 54.7$\pm$0.8 & 53.8$\pm$0.4 & 80.0$\pm$0.4 & 52.4$\pm$0.6 & 79.5$\pm$0.5\\
    & \grey{Supervised oracle} & \grey{60.2$\pm$0.8} & \grey{83.3$\pm$0.9} & \grey{73.3$\pm$0.1} & \grey{92.5$\pm$0.3} & \grey{73.8$\pm$0.3} & \grey{92.4$\pm$0.3}\\
    \midrule
    \multirow{6}{*}{\rotatebox[origin=c]{90}{${U}_{in}$}} & Pseudo-Label~\cite{lee2013pseudo} & 19.4$\pm$0.4 & 43.2$\pm$1.5 & 51.5$\pm$1.2 & 81.2$\pm$0.2 & 49.5$\pm$0.4 & 78.5$\pm$0.2 \\
    & Curriculum Pseudo-Label~\cite{cascante2020curriculum} & 31.4$\pm$0.6 &  55.0$\pm$0.6 & 53.7$\pm$0.2 & 80.2$\pm$0.1 & 53.3$\pm$0.5 & 80.0$\pm$0.5\\
    & FixMatch~\cite{sohn2020fixmatch} & 32.2$\pm$1.0 & 57.0$\pm$1.2 & 56.3$\pm$0.5 & 80.4$\pm$0.5 & \best{58.7$\pm$0.7} & {81.7$\pm$0.2} \\
    & Self-Training & 32.7$\pm$0.2 & 56.9$\pm$0.2 & 56.9$\pm$0.3 & 81.7$\pm$0.2 & {55.7$\pm$0.3} & {82.3$\pm$0.2}\\
    & MoCo~\cite{he2020momentum} & 33.6$\pm$0.2 & 59.4$\pm$0.3 & 55.2$\pm$0.2 & 82.9$\pm$0.2 & 52.5$\pm$0.4 & 79.5$\pm$0.2\\
    & MoCo + Self-Training & \best{39.4$\pm$0.3} & \best{64.4$\pm$0.5} & \best{58.2$\pm$0.5} & \best{84.4$\pm$0.2} & {55.2$\pm$0.5} & \best{82.9$\pm$0.2}\\
    \midrule
    \multirow{6}{*}{\rotatebox[origin=c]{90}{${U}_{in} + {U}_{out}$}} & Pseudo-Label~\cite{lee2013pseudo} & 15.2$\pm$1.0 & 40.6$\pm$1.2 & 52.4$\pm$0.2 & 80.4$\pm$0.5 & 49.9$\pm$0.2 & 78.5$\pm$0.3\\
    & Curriculum Pseudo-Label~\cite{cascante2020curriculum} & 30.8$\pm$0.1 & 54.4$\pm$0.3 & 54.2$\pm$0.2 & 79.9$\pm$0.2 & 53.6$\pm$0.3 & 79.9$\pm$0.2\\
    & FixMatch~\cite{sohn2020fixmatch} & 25.2$\pm$0.3 & 50.2$\pm$0.8 & 51.2$\pm$0.6 & 77.6$\pm$0.3 & 53.1$\pm$0.8 & 79.9$\pm$0.1 \\
    & Self-Training & 32.5$\pm$0.5 & 56.3$\pm$0.3 & \best{55.7$\pm$0.3} & 81.0$\pm$0.2 & \best{55.2$\pm$0.2} & \best{82.0$\pm$0.3}\\
    & MoCo~\cite{he2020momentum} & 44.6$\pm$0.4 & 72.6$\pm$0.5 & 52.9$\pm$0.3 & 81.2$\pm$0.1 & 51.0$\pm$0.2 & 78.5$\pm$0.3 \\
    & MoCo + Self-Training & \best{48.6$\pm$0.3} & \best{74.7$\pm$0.2} & \best{55.9$\pm$0.1} & \best{82.9$\pm$0.2} & 54.0$\pm$0.2 & 81.3$\pm$0.3\\
    \bottomrule
  \end{tabular}
  \caption{\textbf{Results on Semi-Fungi benchmark.} We experiment on Semi-Fungi using the same hyper-parameters from Semi-Aves in Table~\ref{tab:benchmark_aves}. We can see similar conclusions: When training from scratch, MoCo + Self-Training performs the best and adding $U_{out}$ can give an extra performance boost. With expert models, FixMatch and Self-Training (with or without MoCo) is often the best performing one, but the latter is more robust to the out-of-class data.}
  \label{tab:benchmark_fungi}
\end{table*}

\subsection{Implementation details}\label{sec:implementation}
\paragraph{Network architecture and pre-training.} 
For a fair comparison, we use a ResNet-50 network~\cite{he2016identity} on 224$\times$224 images for all our experiments. 
For transfer learning, we use pre-trained models on ImageNet~\cite{ILSVRC15} and iNaturalist 2018 (iNat)~\cite{van2018inaturalist} dataset, which contains 8142 species including 1248 Aves and 321 Fungi species.
Note that there are \emph{no overlapping images} between iNat's training set and Semi-Aves, though there are overlapping categories. 
The images for Semi-Fungi images do not overlap with iNaturalist, but we do not know how many overlapping classes there are as species names were not provided in the original dataset~\cite{fungi_challenge} from which it was constructed.
However, this is less of a concern as we find that iNat pre-trained model performs worse than an ImageNet pre-trained model on Semi-Fungi, suggesting the class overlap is likely small if any.
To obtain an iNat pre-trained model, we train the model using SGD with momentum with a learning rate of 0.0045 and a batch size of 64 for 75 epochs which matches the reported 60\% Top-1 accuracy\footnote{\url{https://github.com/macaodha/inat_comp_2018}}.
We use the ImageNet pre-trained model from torchvision~\cite{pytorch}.

\paragraph{Data augmentation.} For the Semi-Fungi dataset, we first pre-process the images to have a maximum of 300 pixels for each side, while Semi-Aves has a maximum of 500 pixels. We use random-resize-crop to the size of 224$\times$224 and random-flipping for data augmentation, for all the methods except for MoCo and FixMatch. MoCo additionally uses Gaussian blur, color jittering, and random grayscale conversion, while FixMatch uses RandAugment~\cite{cubuk2020randaugment}.

\paragraph{Hyperparameter search.} We found the SSL methods to be sensitive to hyper-parameters such as learning rates, weight decay, etc. As noted in~\cite{oliver2018realistic}, a small validation set poses a risk of picking sub-optimal hyper-parameters. Moreover, labeled data is best used as a source of supervision. While k-fold cross-validation is an alternative, it is expensive. Hence, we use the combined training and validation set for training SSL methods in our experiments and report performance on the test set which is sufficiently large. 
In particular, hyperparameters for all methods were based on the performance on the Semi-Aves dataset and kept fixed for the Semi-Fungi dataset (Tab.~\ref{tab:benchmark_fungi}).
Thus the results in Tab.~\ref{tab:benchmark_aves} should be seen as a validation set performance, while those in Tab.~\ref{tab:benchmark_fungi} represent a novel test set. However, the high-level conclusions are identical across the two benchmarks.

\paragraph{Semi-supervised training.}
 For SSL methods except for FixMatch, we use SGD with a momentum of 0.9 and a cosine learning rate decay schedule~\cite{loshchilov2016sgdr} following~\cite{kornblith2019better,sohn2020fixmatch} for optimization.
 Learning rate and weight decay were picked from a range of [0.03, 0.0001].
 We use a batch size of 64 during training. When there is unlabeled data, we select half of the batch from labeled and another half from unlabeled data (32 each).
 We train models for 10k and 50k iterations for training from expert models and from scratch.
 Other hyper-parameters include threshold $\tau$ for Pseudo-Labeling, which we select from \{0.80, 0.85, 0.90, 0.95\}. 
 When training from scratch, we use $\tau$=0.85 and 0.8 for with and without $U_{out}$; when training from experts we use $\tau$=0.95.
 For Self-Training, we set $T$=1 and $\lambda$=0.7 for all the experiments.
 For FixMatch~\cite{sohn2020fixmatch}, we are able to train the model up to a batch size of 192 (32 labeled and 160 unlabeled images) on 4 GPUs. We find the performance drops significantly with small batch size (\eg 48), however, we are unable to use the same batch size as original paper (\ie 6144) due to limited resources.
 We use a learning rate of 0.01 and threshold $\tau$=0.80 to train FixMatch for 500 epochs when training from scratch and 250 epochs with pre-trained models.
 We report the results from the last training epoch for all the methods. We also notice that FixMatch has more overfitting and the results could be further improved.
 More details are provided in the appendix.
 
\paragraph{Self-supervised training.} 
 We adopt the default settings of MoCo-v2~\cite{chen2020improved}, including MLP projector, 800 training epochs, \etc., but adapt the number of negative samples and learning rate to our task. We use a batch size of 256 and 2048 negative samples in all experiments. We find that using a large number of negative samples (\eg 65,536) hurts the performance.
 When training the MoCo from scratch, we use the default learning rate of 0.03; when training MoCo from ImageNet or iNaturalist pre-trained model, we use a smaller learning rate (0.0003) and fewer training epochs (200) to avoid the potential forgetting problem. In the end, we train a classifier on the global average pooling features of ResNet-50 without freezing the backbone. We find that freezing the feature encoder always leads to worse performance than fine-tuning the entire network.

\subsection{Results}
\begin{figure*}
\centering
\includegraphics[width=0.49\linewidth]{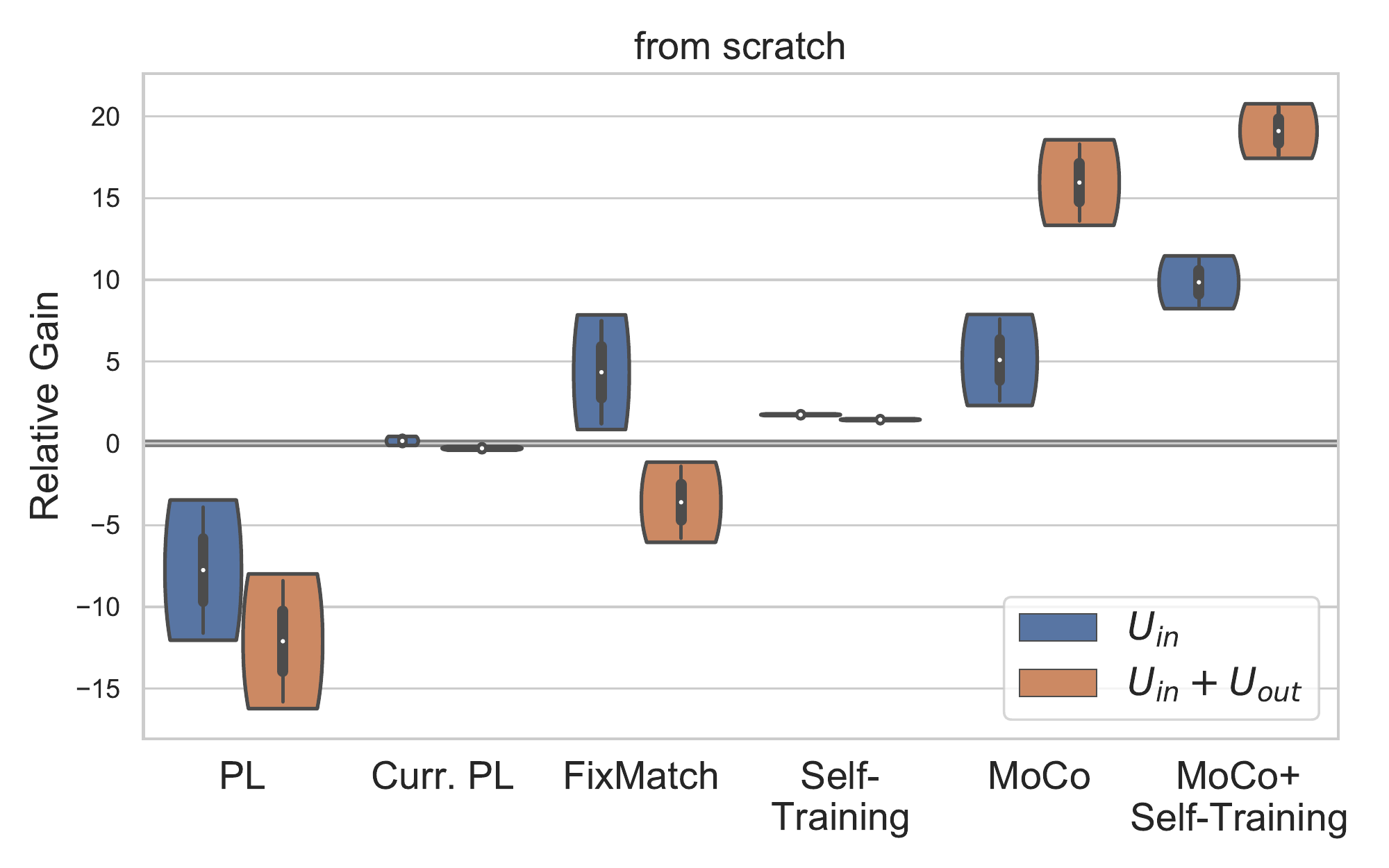}
\includegraphics[width=0.49\linewidth]{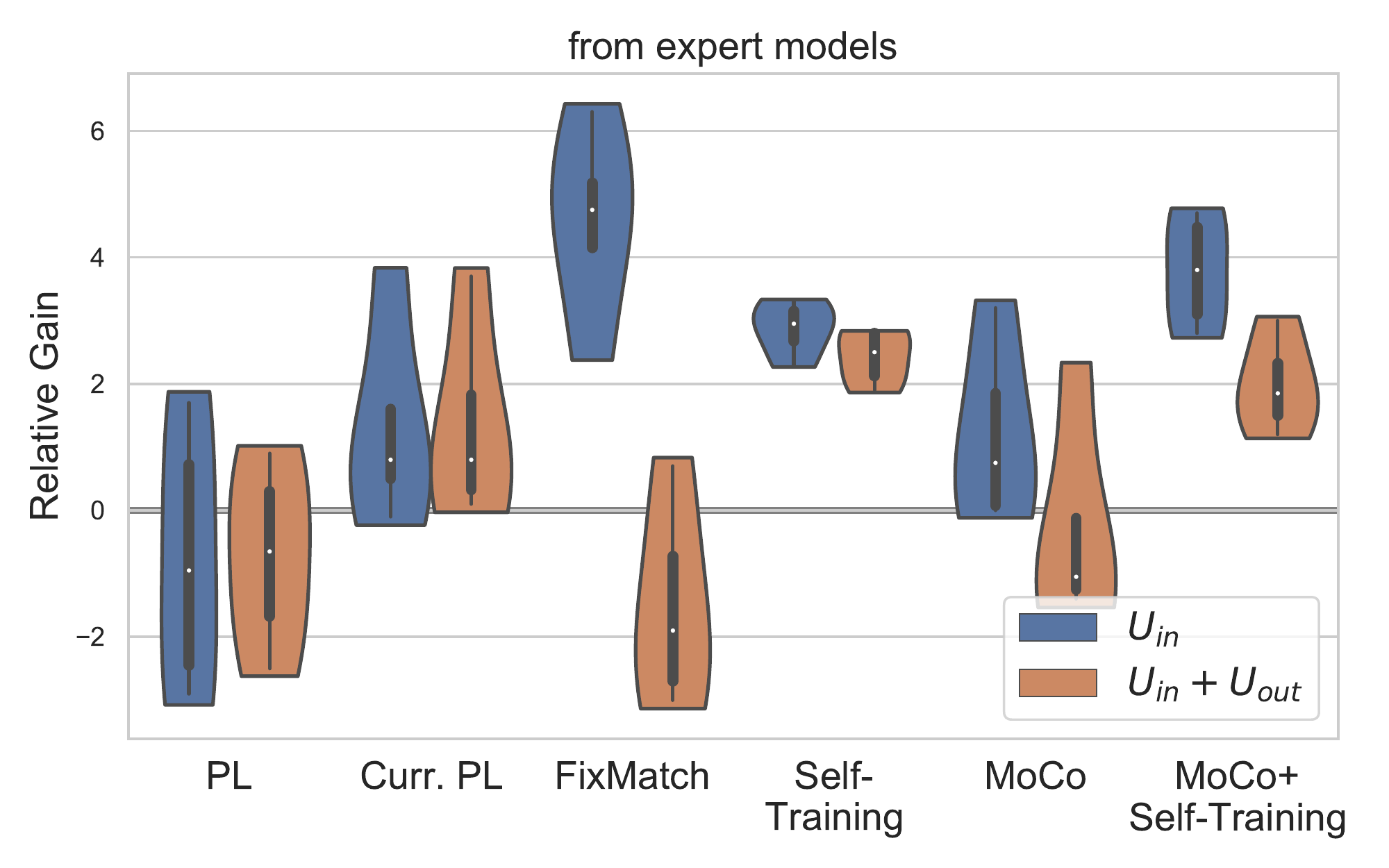}
\caption{\textbf{Relative gains of SSL methods on Semi-Aves and Semi-Fungi.} \textbf{Left:} trained from scratch. \textbf{Right:} using expert models.
For each SSL method, we plot the relative gain, \ie the difference between the supervised baseline in raw accuracy, from the results in both Tab.~\ref{tab:benchmark_aves}~and~\ref{tab:benchmark_fungi}. This shows that (1) the presence of out-of-class data $U_{out}$ often hurts the performance, and (2) Self-Training is often the best method when using pre-trained models.
}
\label{fig:relative}
\end{figure*}

Our experimental results on Semi-Aves and Semi-Fungi are shown in Tab.~\ref{tab:benchmark_aves} and \ref{tab:benchmark_fungi}, respectively. 
To better visualize the results, we plot the relative gain of each SSL method, \ie the differences between supervised baseline in raw accuracy, on both datasets in Fig.~\ref{fig:relative}.
We discuss the results of each setting in the following.

\paragraph{Training from scratch.}
We first discuss the results of training from scratch using only $U_{in}$ on both datasets. 
Comparing to supervised baseline, Curriculum Pseudo-Label does not give improvements and Pseudo-Label even underperforms the baseline. This is possibly due to the low initial accuracy of the model which gets amplified during pseudo labeling.
FixMatch and Self-Training both result in improvements.
Self-supervised learning (MoCo) gives a good initialization and the improvements are similar or even more than using FixMatch. 
Finally, Self-Training using MoCo pre-trained model as the teacher model results in a further 2-3\% improvement.

\paragraph{Using expert models.}
We then consider using an ImageNet or iNat pre-trained model for transfer learning with $U_{in}$ only. 
The transfer learning baseline from either expert model outperforms the best SSL method (MoCo + Self-Training) trained from scratch by a large margin, showing that transfer learning is more powerful in our realistic datasets. 
This observation echos Oliver~\etal~\cite{oliver2018realistic} who showed transferring from ImageNet to CIFAR10 performs better than SSL methods.
Next, we can see that most of the SSL methods, as well as MoCo pre-training, provide improvements over the baselines. The only exception is Pseudo-Label on Semi-Fungi.
Among SSL methods, FixMatch and MoCo + Self-Training perform the best.

\paragraph{Effect of out-of-class unlabeled data.}
Now we consider the setting where the unlabeled data contains both in-class and out-of-class data ($U_{in} + U_{out}$). 
This is the trade-off between more unlabeled data at the cost of a distribution shift.
This effect can be seen in the orange vs.~blue plot in Fig.~\ref{fig:relative}.
When training from scratch, the performances of Pseudo-Label and FixMatch drop by 4-9\%, while Curriculum Pseudo-Label and Self-Training only drop by less than 1\%, showing that they are more robust to the domain shift of unlabeled data.
On the other hand, self-supervised pre-training (MoCo) can benefit significantly from $U_{out}$, providing around 11\% improvement over using $U_{in}$ only on both Aves and Fungi datasets.
Combining with Self-Training gives another 3-6\% improvement, making the gap between transfer learning baseline smaller.

Finally, we consider having $U_{in}+U_{out}$ with expert models.
In Fig.~\ref{fig:relative} we can see the performance often drops in the presence of $U_{out}$.
Curriculum Pseudo-Label and Self-Training are more robust and yield less than 1\% decrease in most cases, while FixMatch is less robust whose performance drops by around 6\%.
The performances of MoCo also drops around 1-3\% and are sometimes worse than the supervised baseline.
Adding Self-Training however provides a 1-3\% boost in performance.
Overall, Self-Training from either a supervised or a self-supervised model is the most robust one.

\paragraph{Robustness to hyper-parameters and trends.} 
We found Pseudo-Label to be sensitive to the threshold $\tau$. When using experts higher thresholds worked better. Increasing the threshold also increased the robustness in the presence of novel classes.
Curriculum Pseudo-Label was found to be more robust in our benchmark, even when adding $U_{out}$.
Self-Training was the most robust to hyper-parameters, we chose the same temperature $T$ and weight $\lambda$ for all the experiments and it consistently improved results regardless of using an expert model or using out-of-domain data.

%% file: conclusion.tex
\section{Conclusion}
There has been a significant interest in self-supervised and semi-supervised learning towards the goal of learning from a few examples. 
However, these methods should be studied in the broader context of approaches for transfer learning, model selection, active learning, and hyperparameter optimization for it to have an impact on realistic applications.
Our benchmark is a step in this direction where we find the strong performance on benchmarks like CIFAR and ImageNet does not always translate to other datasets that violate assumptions implicit in the learning methods.

Self-supervised learning followed by Self-Training is a strong baseline in the absence of experts. 
Of surprise is the marginal gains some SSL methods provide when experts are available. 
Of encouragement is that the simple baseline of Self-Training from experts is robust to out-of-domain data.
Moreover, no method was able to reliably use a large number of out-of-class examples in either domain despite our extensive search over model hyper-parameters and small domain shifts.
Yet, the performances of these methods are far from saturated as indicated by the supervised oracle leaving much room for improvement. 
We hope our proposed benchmarks and results lead to new innovations in SSL.

%% file: supp_arXiv.tex
\section{Appendix}
\setcounter{table}{0}
\setcounter{figure}{0}
\counterwithin{table}{section}
\counterwithin{figure}{section}
\subsection{Results on Semi-Aves with a different split}
Since the original split of Semi-Aves was used for hyper-parameter selection, we create another split of the dataset for additional evaluation, which can be seen as a two-fold cross-validation.
We first merge the images from $L_{in}$, $U_{in}$, and test sets, then randomly redistribute. The classes and the number of images of each class are kept the same in each set.
The $U_{out}$ set is also unchanged. 
Thus this split has roughly the same difficulty as the original split but contains new in-domain training and test images.
We show the results of this split in Tab.~\ref{tab:benchmark_aves_split}, using the same hyper-parameters of the main paper.
The trends are similar --- Self-Training and FixMatch are both effective, but FixMatch is affected negatively by the out-of-class data.


\subsection{SSL benchmark on the CUB dataset}
Since Semi-Aves and Semi-Fungi are new datasets, here we provide another benchmark based on the widely-used Caltech-UCSD Birds-200-2011 (CUB) dataset~\cite{WahCUB_200_2011}.
The original class labels are sorted by the species name.
Hence, we select the 100 \emph{odd} classes as in-class species and 100 \emph{even} classes as out-of-class species, to ensure a low domain mismatch between $U_{in}$ and $U_{out}$.
There are 41-60 images per class originally (disregard the original training/test split).
For each in-class species, we select 5/5/10 images for $L_{in}$, validation set, and test set. The rest (21-40 images) are used for unlabeled data $U_{in}$.
For out-of-class species, all the images are included in $U_{out}$.
The statistics of the dataset split is shown in Tab.~\ref{tab:cub}.
Since the dataset is quite small, we did not use self-supervised learning (MoCo) for pre-training, and we use the validation set to select the best model. 
The results on the CUB dataset are shown in Tab.~\ref{tab:benchmark_cub}. 
In this benchmark, we can see that both Curriculum Pseudo-Label and Self-Training are helpful, even when having $U_{out}$ included. This is potentially due to the small domain mismatch between the two sets of unlabeled data.

\begin{table}[h!]
\setlength{\tabcolsep}{8pt}
\centering    
  \begin{tabular}{c c c c c c }
  \toprule
  split $\rightarrow$& $L_{in}$ & val & test & $U_{in}$ & $U_{out}$\\
  \midrule
  \#images $\rightarrow$& 500 & 500 & 1000 & 3853 & 5903\\
  \bottomrule
  \end{tabular}
  \caption{\textbf{Number of images in the CUB benchmark.}}
  \label{tab:cub}
\end{table}

\definecolor{mygray}{gray}{0.7}
\begin{table*}[th!]
  \setlength{\tabcolsep}{15pt}
  \centering
  \begin{tabular}{c c | c c | c c | c c}
    \toprule
    &\multirow{2}{*}{Method} & \multicolumn{2}{c|}{\textbf{from scratch}} & \multicolumn{2}{c|}{\textbf{from ImageNet}} & \multicolumn{2}{c}{\textbf{from iNat}}\\
    && Top1 & Top5 & Top1 & Top5 & Top1 & Top5 \\
    \midrule
    & Supervised baseline & 21.8&	42.9&	51.9&	76.0&	66.7&	85.9\\
    &\grey{Supervised oracle} & \grey{56.0} & \grey{78.7} & \grey{71.9} & \grey{89.4} & \grey{76.7} & \grey{91.2}\\
    \midrule
    \multirow{6}{*}{\rotatebox[origin=c]{90}{${U}_{in}$}} & Pseudo-Label~\cite{lee2013pseudo} & 18.0&	37.4&	54.3&	79.1&	66.3&	86.4\\
    & Curriculum Pseudo-Label~\cite{cascante2020curriculum} & 20.0&	41.5&	53.4&	79.0&	70.0& \best{88.7}\\
    & FixMatch~\cite{sohn2020fixmatch} & 24.0	& 47.6 & \best{58.2} & 79.6 &	70.4 & 88.2\\
    & Self-Training & 23.7&	45.1&	53.9&	76.8&	67.6&	86.4 \\
    & MoCo ~\cite{he2020momentum} & 28.5 & 54.7 & 52.8 & {79.2} & 69.4 & {88.3} \\
    & MoCo + Self-Training & \best{34.0} & \best{58.9} & {56.6} & \best{80.1} & \best{71.1} & {88.3}\\
    \midrule
    \multirow{6}{*}{\rotatebox[origin=c]{90}{${U}_{in}+{U}_{out}$}} & Pseudo-Label~\cite{lee2013pseudo} & 11.8&	30.7&	53.6&	78.1&	66.8&	86.4 \\
    & Curriculum Pseudo-Label~\cite{cascante2020curriculum} & 21.3 &	42.1&	53.8& {79.4}&	\best{69.9}&	\best{88.5}\\
    & FixMatch~\cite{sohn2020fixmatch} & 17.5 & 39.7 & 50.8 & 74.4 &	65.1 & 85.1\\
    & Self-Training & 23.0&	45.0&	{54.3}&	77.1&	68.0&	86.1\\
    & MoCo~\cite{he2020momentum} & 37.9 & 65.4 & 51.0 & 78.6 & 68.5 & {87.8} \\
    & MoCo + Self-Training & \best{40.8} & \best{66.8} & \best{55.0} & \best{80.2} & {68.9} & {87.9}\\
    \bottomrule
  \end{tabular}
  \caption{\textbf{Results on Semi-Aves benchmark with a different split.} Using the same hyper-parameters, we can see similar conclusions here as using the original split. Overall, Self-Training and FixMatch are effective but out-of-class data often hurts the performance.
  }
  \label{tab:benchmark_aves_split}
\end{table*}

\begin{table*}[th!]
  \setlength{\tabcolsep}{15pt}
  \centering    
  \begin{tabular}{c c | c c | c c | c c}
    \toprule
    &\multirow{2}{*}{Method} & \multicolumn{2}{c|}{\textbf{from scratch}} & \multicolumn{2}{c|}{\textbf{from ImageNet}} & \multicolumn{2}{c}{\textbf{from iNat}}\\
    && Top1 & Top5 & Top1 & Top5 & Top1 & Top5 \\
    \midrule
    & Supervised baseline & 11.1 & 27.4 & 58.7 & 85.8 & 77.3 & 94.2\\
    &\grey{Supervised oracle} & \grey{68.5} & \grey{87.8} & \grey{84.5} & \grey{97.1} & \grey{90.0} & \grey{98.0}\\
    \midrule
    \multirow{4}{*}{\rotatebox[origin=c]{90}{${U}_{in}$}} & Pseudo-Label~\cite{lee2013pseudo} & 13.5 & \best{32.1} & 57.0 & 85.3 & 78.3 & 95.7\\
    & Curriculum Pseudo-Label~\cite{cascante2020curriculum} & \best{14.5} & 31.4 & 57.3 & 84.7 & 80.1 & \best{96.7} \\
    & {FixMatch}~\cite{sohn2020fixmatch} & 10.7 & 26.6 & 53.2 & 79.8 & \best{81.6} & 95.2\\
    & Self-Training & 12.6 & 29.8 & \best{61.3} & \best{86.3} & 80.6 & 95.8\\
    \midrule
    \multirow{4}{*}{\rotatebox[origin=c]{90}{${U}_{in}+{U}_{out}$}} & Pseudo-Label~\cite{lee2013pseudo} & 11.9 & 30.8 & 59.1 & 86.1 & 77.7 & 94.8\\
    & Curriculum Pseudo-Label~\cite{cascante2020curriculum} & \best{12.9} & \best{32.3} & {59.6} & \best{86.5} & \best{81.2} & \best{96.8} \\
    & {FixMatch}~\cite{sohn2020fixmatch}  & 10.7 & 27.1 & 52.8 & 81.7 & 78.6 & 95.7\\
    & Self-Training & 12.2 & 29.2 & \best{61.4} & 85.9 & 79.9 & 96.0\\
    \bottomrule
  \end{tabular}
  \caption{\textbf{SSL benchmark on the CUB dataset.} In this benchmark, we can see both Curriculum Pseudo-Label and Self-Training are helpful, even with out-of-class unlabeled data.} 
  \label{tab:benchmark_cub}
\end{table*}

\subsection{Related prior work on SSL analysis}
\textbf{On out-of-class unlabeled data.}
Oliver \etal\cite{oliver2018realistic} showed that out-of-class unlabeled data negatively impacts performance, but analysis was done on CIFAR-10 with images from 6 labeled and 4 unlabeled classes. The classes are quite different making the problem of selecting in-domain images relatively easy in comparison to fine-grained domains --- in our benchmarks the out-of-class data $U_{out}$ are other species of birds or fungi.
In fact, we show that more out-of-class data helps when using self-supervised and self-training methods trained from scratch. 
However, the additional data does not seem to help when initialized with experts.

\textbf{On transfer learning.}
Oliver \etal showed a transfer learning accuracy of 87.9\% on CIFAR-10 with 4k labels, outperforming many SSL methods including PL~\cite{lee2013pseudo} and VAT+EM~\cite{miyato2018virtual}. Although recent results are better, the low resolution of CIFAR-10 (32$\times$32 pixels) makes transfer learning from ImageNet less effective.
{On STL-10 that has a higher resolution (96$\times$96 pixels), fine-tuning a ImageNet pre-trained ResNet-50 model on 5k labels provides 97.2\% accuracy, while that trained on iNaturalist provides 95.0\% accuracy. This beats 94.8\% of FixMatch using 5k labeled examples when trained from scratch. Note that the iNaturalist dataset has no overlap with STL-10, yet transfer learning is effective.}

\begin{figure}[t!]
\centering
\includegraphics[width=0.99\linewidth]{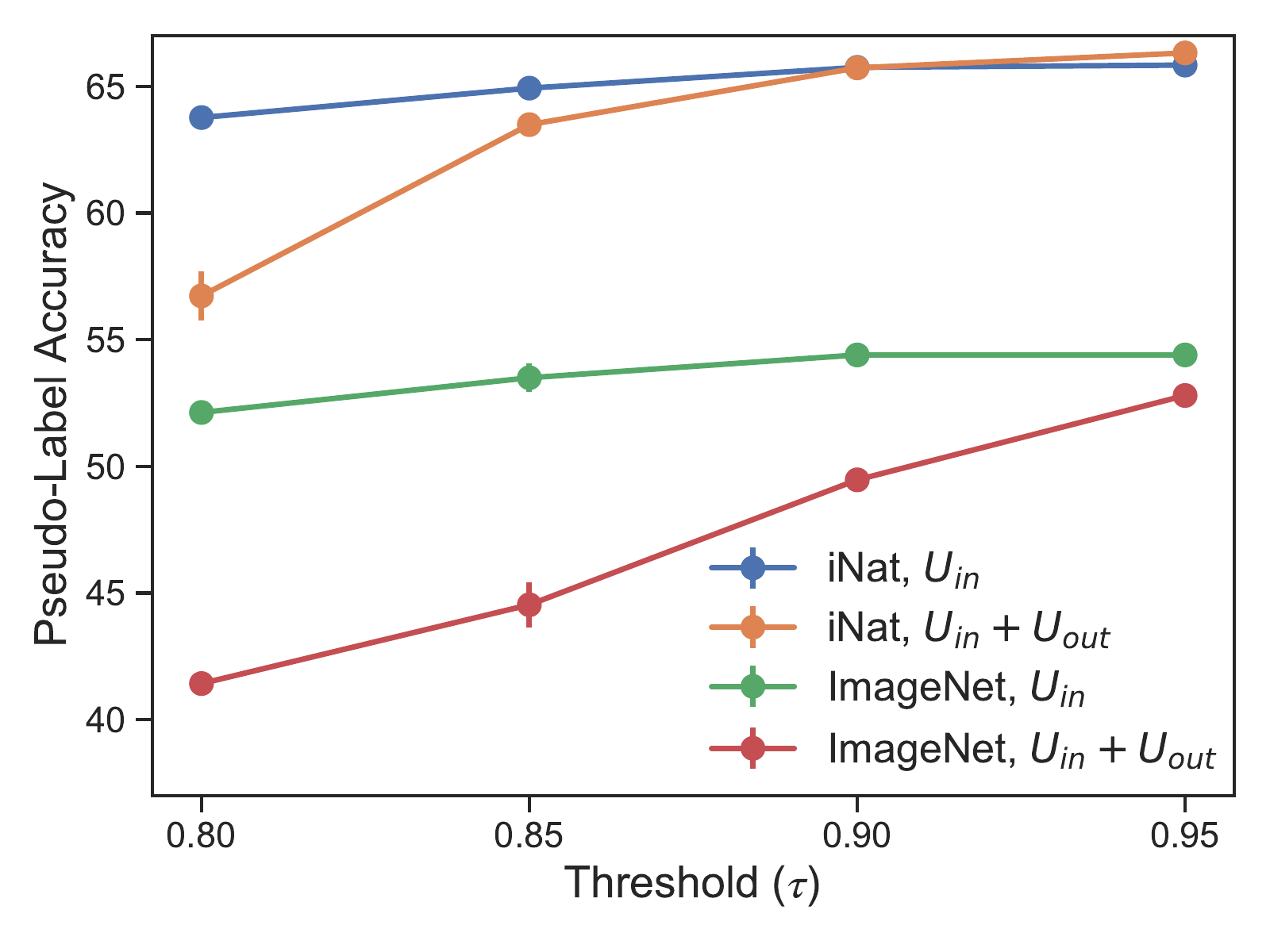}
\caption{\textbf{Pseudo-label with different threshold $\tau$.} Pseudo-label is sensitive to the threshold. The negative impact of out-of-class unlabeled data is reduced by increasing the threshold, yet when the initial performance is low the scheme is not effective as seen by the performance of the ImageNet pre-trained model.}
\label{fig:threshold}
\end{figure}

\begin{figure*}[ht!]
\centering
\includegraphics[width=0.33\linewidth]{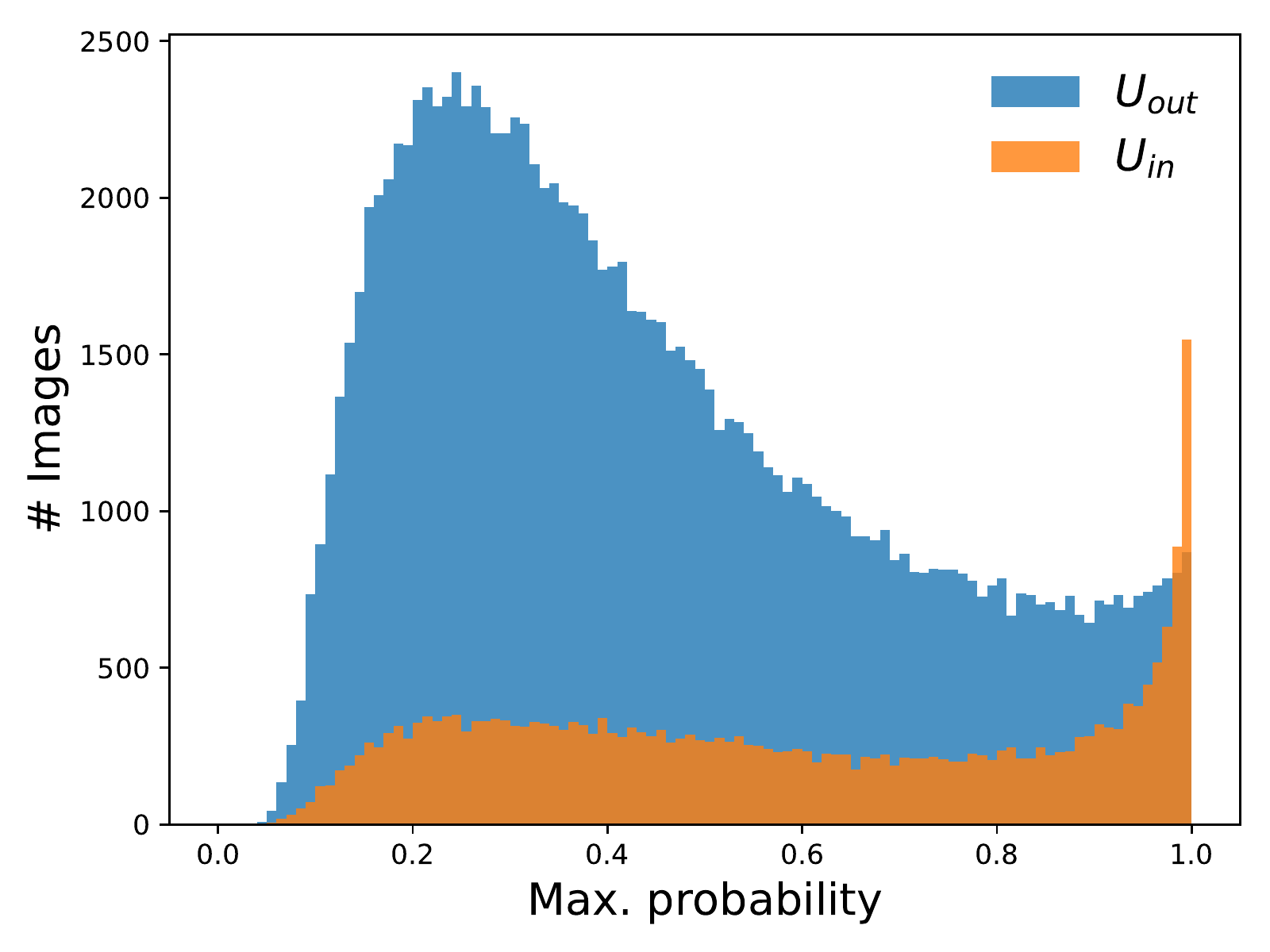}
\includegraphics[width=0.33\linewidth]{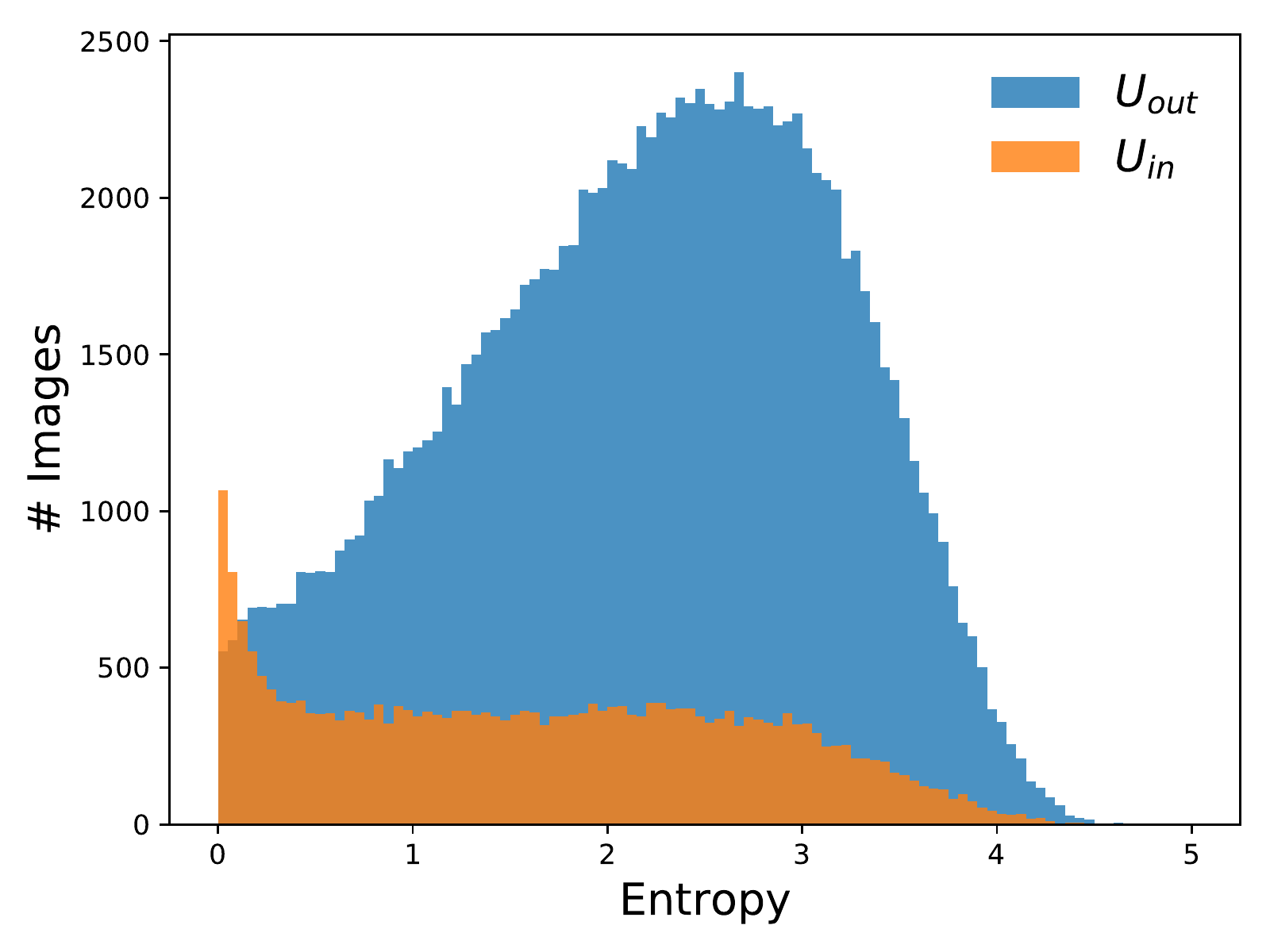}
\includegraphics[width=0.33\linewidth]{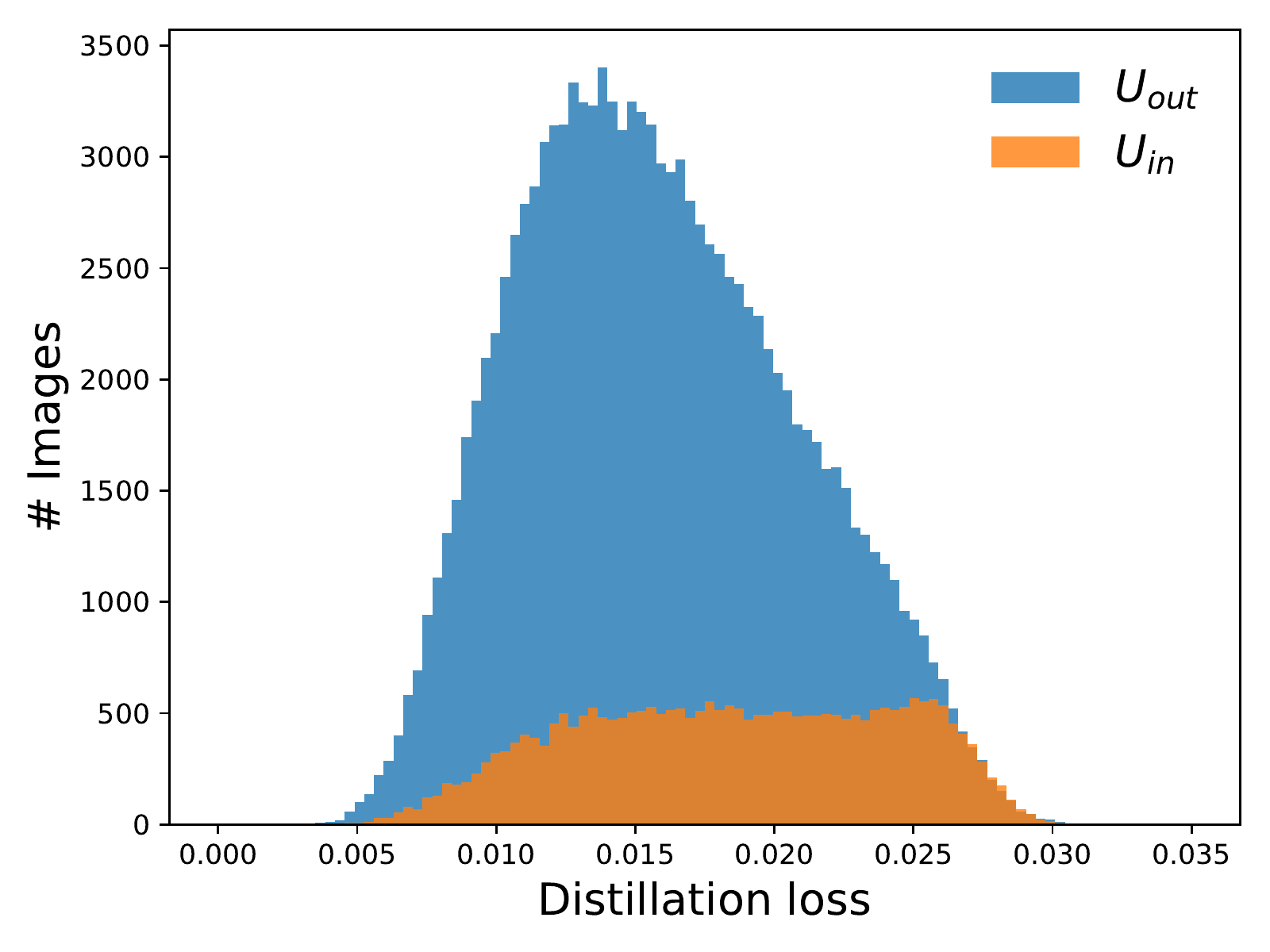}
\caption{\textbf{Predictions of unlabeled data using a supervised model.} 
We plot the distribution of the predictions of data from $U_{in}$ and $U_{out}$.
Specifically, we plot the maximum probability of the class predictions (left), entropy of the predictions (middle), and the distillation loss between the teacher and student model before the training starts (right).
Unlabeled data from the same distribution tend to have a higher maximum probability, a lower entropy, or a higher distillation loss.
}
\label{fig:entropy}
\end{figure*}

\subsection{Analysis on out-of-class unlabeled data}
\textbf{The effect of threshold parameter for Pseudo-Label.}
We found Pseudo-Label method is sensitive to the threshold parameter $\tau$. 
Fig.~\ref{fig:threshold} plots the accuracy as a function of $\tau$ with different unlabeled data and experts on Semi-Aves. 
A higher threshold performs better, especially in the presence of out-of-class data $U_{out}$ as this excludes novel class images where the confidence of prediction is likely to be low.
On the other hand, lower values work just as well when unlabeled data is in-domain $U_{in}$.
However, this scheme only appears to work when using strong experts (\eg, iNat) whose confidence is likely calibrated, unlike random or ImageNet pre-trained model, where the presence of out-of-class data reduces performance. 
This poses a practical problem for this method --- increasing the threshold increases robustness but reduces the amount of unlabeled data that is used during training.

\subsection{The effect of out-of-class unlabeled data.}
To see how the domain mismatch between $U_{in}$ and $U_{out}$ can affect SSL methods, we analyze the predictions of the unlabeled data.
We use the supervised model trained on $L_{in}$ to compute the predictions of the unlabeled data on the Semi-Aves dataset. 
We plot the histogram of the maximum probability and the entropy of the predictions of $U_{in}$ and $U_{out}$ in Fig.~\ref{fig:entropy} (left and middle). 
We also plot the distribution of the distillation loss, which is calculated between the supervised model (teacher) and the ImageNet pre-trained model (student), with a temperature $T=1$ (Fig.~\ref{fig:entropy} right).
This is in the beginning of the self-training process and the last layer of the student model is randomly initialized.
Overall, the model is generally more uncertain about the out-of-class data, which often has a higher entropy or a smaller maximum probability.
The distillation loss on $U_{in}$ is also often higher than that of $U_{out}$, suggesting the model focuses more on those from $U_{in}$ during training.
However, there is still a good amount of data from $U_{out}$ having a high maximum probability, which has a negative impact for pseudo-label methods.

\subsection{Implementation details of FixMatch}
For FixMatch, we used the official Tensorflow code and a PyTorch re-implementation for our experiments.
The PyTorch version did reproduce the results on CIFAR-10 with 4000 labels (95.68\% vs 95.74\% reported in the paper).
We found the optimization details such as learning rate, batch size, and number of epochs to be crucial for FixMatch. 
Due to the limitation of our resources, we can only use a batch size of 64 for labeled data (and 320 for unlabeled data), where the original paper used a batch size of 1024 (and 5120 for unlabeled data). 
Since there is small discrepancy between the two implementations, we reported the best result among the two for each setting: Tensorflow version for training from scratch and PyTorch version when using expert models.

\subsection{Value for computing}
Among the SSL methods, Pseudo-Label requires the least amount of computation, but it does not uniformly lead to improvements in our benchmark.
Curriculum Pseudo-Label trains a model several times (six in our implementation), hence is more expensive, though the performance saturates after the first few iterations.
FixMatch requires more training epochs and is the most time-consuming comparing to other SSL methods, but the performance is the best when having expert model with in-class unlabeled data only.
Self-training only needs two rounds of training, one for training the teacher model and one for the student. However, it is often the best and is more robust to out-of-class data.